\documentclass[letterpaper, 10 pt, conference]{ieeeconf}  
\IEEEoverridecommandlockouts                              
\usepackage{graphics} 
\usepackage{color}
\usepackage{epsfig} 

\usepackage{multirow}
\usepackage{comment}
\usepackage{afterpage}
\usepackage{setspace}
\usepackage{amsmath}
\usepackage{algorithm}
\usepackage{overpic}
\usepackage[noend]{algorithmic}

\DeclareMathOperator*{\argmax}{arg\,max}

\title{\LARGE \bf
Annotation Cost Reduction of Stream-based Active Learning \\
by Automated Weak Labeling using a Robot Arm
}

\author{Kanata Suzuki, Taro Sunagawa, Tomotake Sasaki, and Takashi Katoh
\thanks{Kanata Suzuki, Taro Sunagawa, Tomotake Sasaki, and Takashi Katoh are with Artificial Intelligence Laboratories, Fujitsu Laboratories LTD., Kanagawa 211-8588, Japan. E-mail: {\tt\small suzuki.kanata@fujitsu.jp}.}}

\begin{document}
\maketitle
\thispagestyle{empty}
\pagestyle{empty}


\begin{abstract}
Stream-based active learning (AL) is an efficient training data collection method, and it is used to reduce human annotation cost required in machine learning. However, it is difficult to say that the human cost is low enough because most previous studies have assumed that an oracle is a human with domain knowledge. In this study, we propose a method to replace a part of the oracle's work in stream-based AL by self-training with weak labeling using a robot arm. A camera attached to a robot arm takes a series of image data related to a streamed object, which should have the same label. We use this information as a weak label to connect a pseudo-label (estimated class label) and a target instance. Our method selects two data from a series of image data; high confidence data for correcting pseudo-labels and low confidence data for improving the performance of the classifier. We paired a pseudo-label provided to high confidence data with a target instance (low confidence data). By using this technique, we mitigate the inefficiency in self-training, that is, difficulty in creating pseudo-labeled training data with a high impact on the target classifier. In the experiments, we employed the proposed method in the classification task of objects on a belt conveyor. We evaluated the performance against human cost on multiple scenarios considering the temporal variation of data. The proposed method achieves the same or better performance as the conventional methods while reducing human cost.
\end{abstract}

\section{Introduction}
\label{sec:intro}

\subsection{Background}
The performance of image recognition tasks has been significantly improved by modern machine learning methods represented by deep learning~\cite{dl1}. However, many of these methods require large amounts of labeled training data~\cite{dl2}; therefore, we need a high human annotation cost to apply them to real-world problems~\cite{an1}\cite{an2}\cite{an3}. In particular, we need a high human annotation cost for applications when we deal with a temporal variation of target data, because we should make data depending on the new situations~\cite{online1}\cite{online2}\cite{online3}. For instance, considering the product classification in factory lines or logistics, the distribution of target products may change frequently. In this paper, we study the human annotation cost reduction using a robot arm described in the next subsection. 

Active learning (AL)~\cite{ent} is a strategy to reduce human annotation cost.
It requires an ``oracle" to label some unlabeled data that the classifier, such as a deep neural network (DNN), has low confidence about its estimation for the class~\cite{al2}\cite{al3}\cite{al4}. However, it is difficult to say that human annotation cost is low enough because most previous studies assume that an oracle is a human with domain knowledge. In addition, considering the temporal variation of data, stream-based AL for sequentially provided data is more suitable in adaptability than pool-based AL that is for accumulated data. However, the stream-based AL requires uninterrupted supervision that is costly for a human.

\setlength\textfloatsep{5pt}
\begin{figure}[t]
  \centering
  \includegraphics[width=8.6cm]{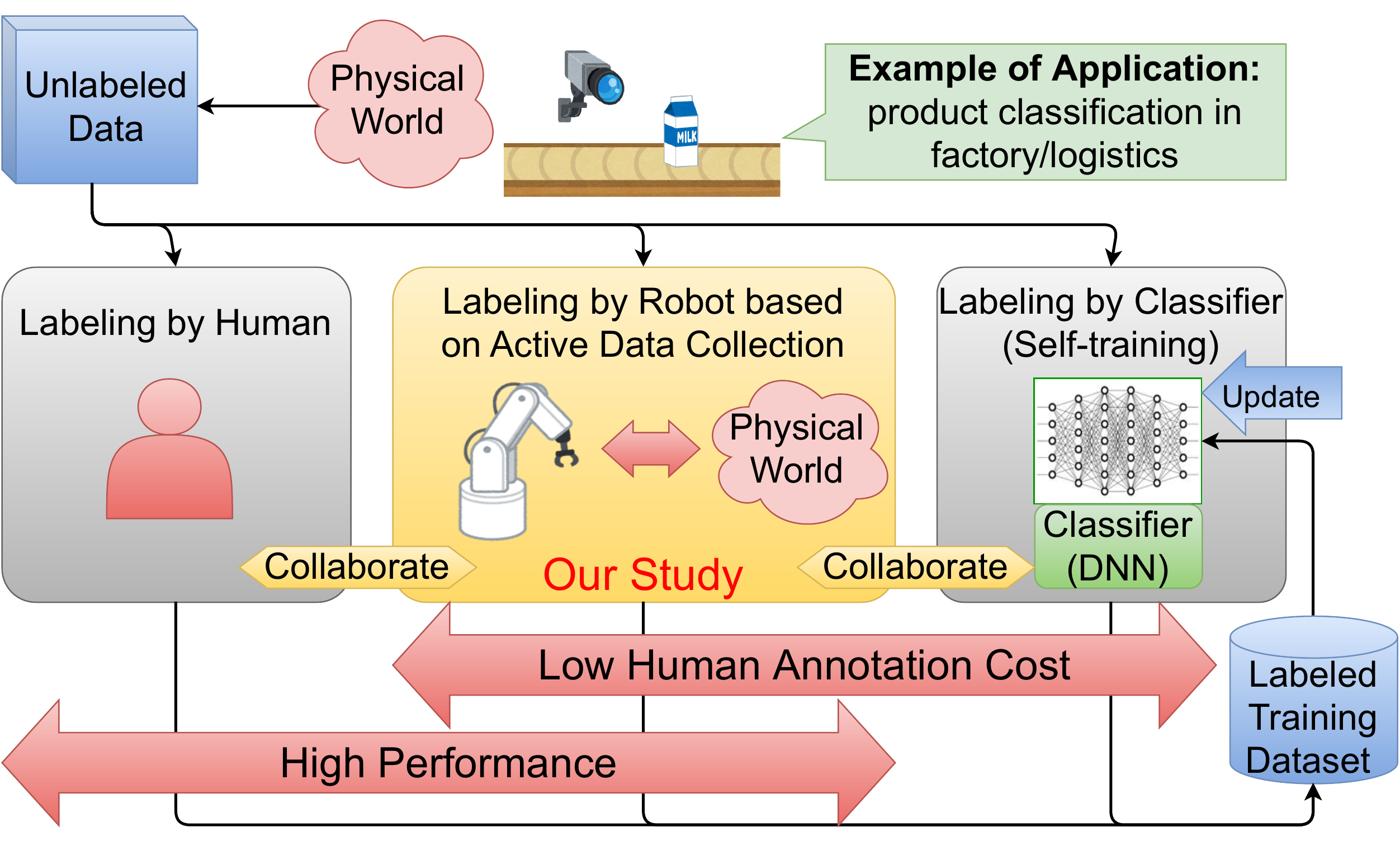}
  \caption{
The concept of our study: The labeling by a human greatly improves the performance of the classifier than the labeling by the classifier itself. However, it requires a large human annotation cost. To reduce the cost, the robot arm collects weakly labeled data from the physical world for enhancing self-labeling.
  }
  \label{fig:intro}
\end{figure}

Unlabeled data usage technologies such as semi-supervised~\cite{semi1}\cite{semia2}, representation~\cite{semi2}\cite{rep1}, and transductive~\cite{domein1}\cite{trans1} learning are other strategies to reduce human annotation cost. Self-training is one of the semi-supervised learning methods, which provides estimated labels (pseudo-labels) for unlabeled data by using the classifier in training itself. Most supervised learning algorithms can be extended to semi-supervised ones by self-training because it is a wrapper that does not depend on the base method. Wrong pseudo-labels, however, worsen the performance of the classifier. To prevent this, the self-training calculates the confidence of a pseudo-label and uses only the high confidence data to re-train. However, the data the classifier has high confidence are also the data with little impact on the training of classifier.

Auxiliary classifiers can be used for pseudo-labeling in addition to the target classifier itself to avoid the inefficiency described above~\cite{co1}\cite{co2}\cite{co3}. To make such ensemble pseudo-labeling by these classifiers working effectively, the auxiliary classifier should label with high confidence for some data by an estimation process that is somewhat different from the main classifier's process. Unfortunately, it is difficult to obtain such an ideal auxiliary classifier in general. Notably, a human oracle in AL can be regarded as a highly ideal auxiliary classifier in self-training if without considering human annotation cost. 

\subsection{Proposal}
In this study, we propose a Robot-Assisted AL to replace a part of the oracle's work by self-training with weak labeling. A camera attached to the robot arm takes a series of image data related to a new image data. The labels of these data are unknown at this stage, but it can be assumed that they have the same label, and we call the data characterized by this property weakly labeled data. We use the weak label to connect pseudo-label and target instance. Our method selects two data from a series of image data; one has high confidence and it can be expected to correct pseudo-labeling, another one has low confidence and it can be expected to improve the performance of the classifier. We paired a pseudo-label provided to high confidence data with a target instance (low confidence data). By using this mechanism, we aim to solve the inefficiency problem in self-training, that is, the difficulty in creating pseudo-labeled training data with a high impact on the target classifier.

We decided to use stream-based AL due to the time limitation issue that exists in the data collection process in the physical world. To our best knowledge, there are some previous studies for the label acquisition by a robot arm~\cite{r1}\cite{r2}\cite{r3}, but no algorithm for robot-specific stream-based AL; therefore, we use a standard AL algorithm assuming a human oracle. Also in this study, we prepared a separate hard-coded classifier to boost the performance of the main classifier in the early stages of stream-based AL. We assume an identifier (a tool indicating the class label, e.g. barcode) in this study, and we use an identifier detector as an auxiliary classifier on the self-training to automated pseudo-labeling. That is, we regard the identifier detector as a fixed classifier that can label with high confidence only for data containing a clearly captured identifier.

Data that fail to be automatically pseudo-labeled are labeled by a human, and the number of human-labeled data is treated as human annotation cost. Considering the temporal variation of data, we experimentally evaluate the performance against human annotation cost on multiple scenarios. As a result, our method achieves the same or better performance while reducing human annotation costs. The contributions of this research are as follows.
\begin{enumerate}
    \item Proposal of the Robot-Assisted AL that enables efficient self-training by using weakly labeled data and an identifier detector as an auxiliary classifier.
    \item Experimental evaluation of the proposed method by scenarios simulating the image-based object classification task.
\end{enumerate}

\begin{algorithm}[t]
\caption{Training procedure in stream-based AL}
\begin{algorithmic}[1]
    \REQUIRE $\delta_E,\Delta\delta_E,\delta_v,N_{train},N_{max},N_{iter},\eta$
    \WHILE{$n(D)\leq\,N_{max}$}
        \STATE get image $x_o$ from streamed object
        \STATE compute $E(x_o)$ according to Eq. (1)
        \IF{$E(x_o) > \delta_{E}$}
            \STATE get $\{ x_1,x_2,\cdots,x_{N_{sub}} \} \sim X_{sub}$
            \STATE select $x^*$ by using Eq. (2)
            \STATE detect $l_d^*$ from $\{ x_1,x_2,\cdots,x_{N_{sub}} \}$
            \IF{$l_d^* \in \{ 1,2,\cdots,C \}$}
                \STATE add sample: $D \leftarrow D \cup \{(x^*,l_d^*)\}$
            \ELSE
                \STATE compute $v$ and $l_p^*$ by using Eq. (3)
                \IF{$v > \delta_v$}
                    \STATE add sample: $D \leftarrow D \cup \{(x^*,l_p^*)\}$
                \ELSE
                    \STATE add sample: $U \leftarrow U \cup \{x^*\}$
                \ENDIF
            \ENDIF
        \ENDIF
        \IF{$(n(D) \ne 0)$ \AND $(n(D)\%N_{train}=0)$}
            \FOR{$i=1$ to $n(U)/2$}
                \STATE sample $x_U^* \sim U$
                \STATE obtain $l_a$ by human labeling
                \STATE add sample: $D \leftarrow D \cup \{(x_U^*,l_a)\}$
            \ENDFOR
            \STATE initialize parameters: $\theta \leftarrow \theta'$
            \FOR{$i=1$ to $N_{iter}$}
                \STATE train and update model: $\theta \leftarrow \theta - \eta \nabla L$
            \ENDFOR
            \STATE update threshold: $\delta_E \leftarrow \delta_E + \Delta\delta_E$
        \ENDIF
    \ENDWHILE
\end{algorithmic}
\label{alg:AL}
\end{algorithm}

\setlength\textfloatsep{5pt}
\begin{figure*}[t]
  \centering
  \includegraphics[width=17.6cm]{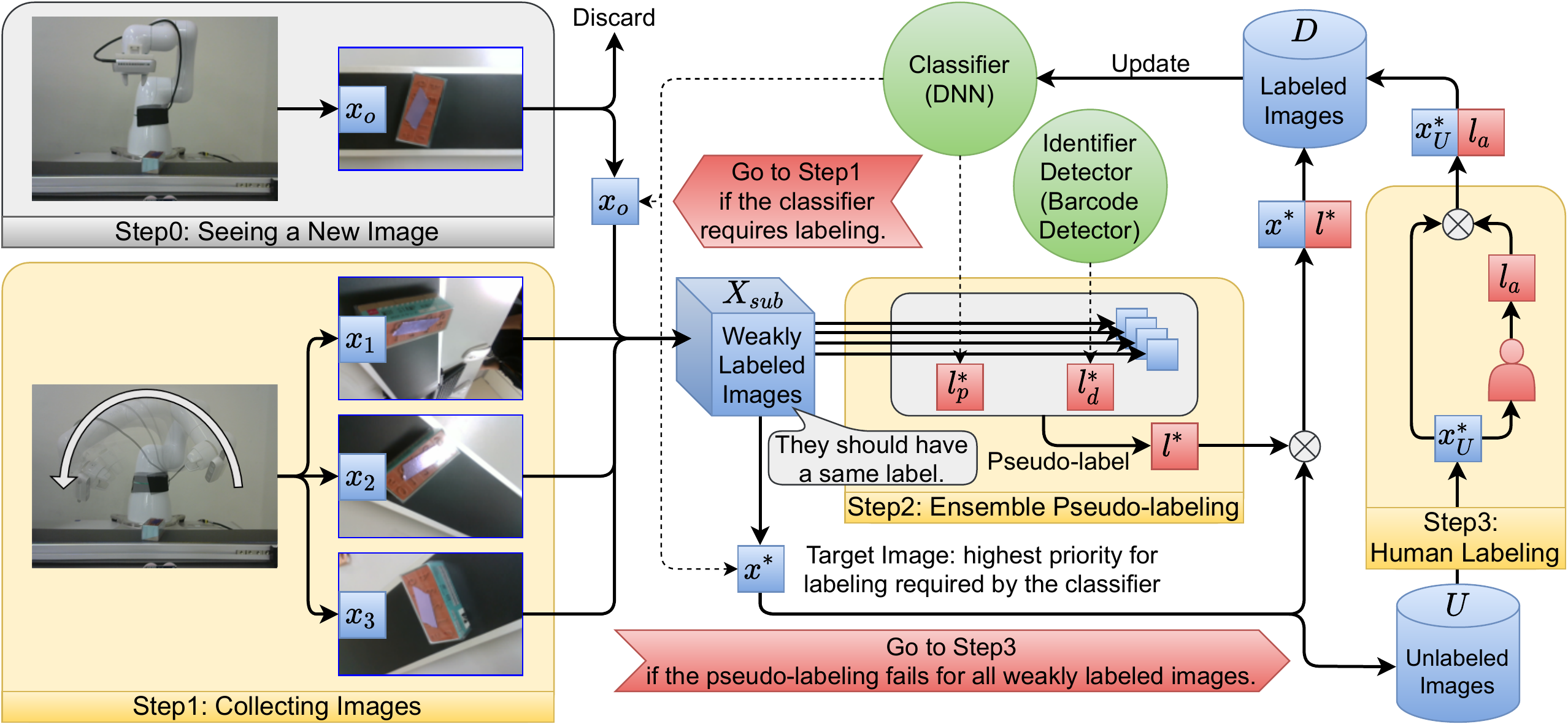}
  \caption{
Overview of the proposed Robot-Assisted AL. The robot captures the image of the top of the objects and selects whether it is effective data for the training of the classifier (step 0). If the classifier requires labeling, the robot arm captures a series of image data $X_{sub}$ (step 1). The method performs ensemble pseudo-labeling (step 2); the identifier detector assigns pseudo-label $l_d^*$ from detected identifier from $X_{sub}$, or the classifier assigns a pseudo-label $l_p^*$. A pseudo-label $l^*$ determined from two paths and the target image $x^*$ determined from $X_{sub}$ are added to the training dataset $D$. Human annotations are performed only on the dataset $U$ that cannot be assigned pseudo-labels (step 3).
  }
  \label{fig:method}
\end{figure*}

\section{Robot-Assisted Active Learning}
\label{sec:method}
Fig.~\ref{fig:method} shows an overview of the proposed Robot-Assisted AL. In this section, we first present the whole procedure of our method in stream-based AL with a concrete task~(Sec.~\ref{sec:AL}). Then, we introduce the individual functions; query strategy, ensemble pseudo-labeling, and self-training with pseudo-labels~(Sec.~\ref{sec:query} to \ref{sec:update}).

\subsection{Procedure of our method in Stream-based AL}
\label{sec:AL}
Stream-based AL is one of the major settings of AL. In stream-based AL, each unlabeled instance is supplied from the data source, and the learner queries if the unlabeled instance is effective in learning~\cite{ent}. Algorithm~\ref{alg:AL} and Fig.~\ref{fig:method} show the training procedure of the proposed method. The proposed method consists of a classifier to classify the object, an identifier detector as the auxiliary classifier, and a robot system to collect data from the physical world. In our task, a robot arm is placed in front of a belt conveyor, which carries the objects one by one periodically. The object is assumed to have a barcode as an identifier that has the information of the object class, and we consider a barcode reader as the identifier detector. Note that, our task settings are experimental, and the barcode can be replaced with other identifiers, such as a tag, printed logo or mark, embossed/carved model number, etc. in a real problem setting.

The robot arm captures an image $x_o$ of an object from above and an image classifier outputs entropy $E(x_o)$. The system decides whether to label or discard the image of the streamed object on the basis of the query strategy (Sec.~\ref{sec:query}).

If the classifier requires labeling, the robot arm captures a series of image data $X_{sub}$ while changing the position of its arm tip. The images in $X_{sub}$ should have the same class label. We use this information as a weak label. We call $X_{sub}$ weakly labeled images. Using $X_{sub}$, the system performs ensemble pseudo-labeling (Sec.~\ref{sec:psuedo}) and assigns pseudo-label $l^*$ to the target image $x^*$, for which the classifier requires labeling with the highest priority. In that process, we use two types of pseudo-labeling with the classifier and the identifier detector that estimates the object class by detecting the barcode. If the system cannot give a pseudo-label, the target image is added to the unlabeled dataset $U$.

The classifier is regularly trained with the obtained dataset $D$ through the stream-based AL. Parameters $\theta$ of the classifier are updated by the training every time $N_{train}$ images are added to $D$. Before the training, the system adds randomly sampled data from $U$ to $D$ with human labeling (Sec.~\ref{sec:human}). The parameters $\theta$ are initialized with $\theta'$ obtained by a pre-training, and they are updated $N_{iter}$ times. The system repeats the above procedure until the number of data in $D$ becomes $N_{max}$.

\subsection{Query strategy}
\label{sec:query}
We use uncertainty sampling~\cite{ent} for the query strategy of our method. As a measure of the uncertainty, we employ the Shannon entropy calculated as follows:
\begin{eqnarray}
  E(x) = -\sum_{c=1}^C f_{\theta}(x)_c \log f_{\theta}(x)_c,
  \label{eq:entropy}
\end{eqnarray}
where $x$ is an input image, $f_{\theta}(\cdot)_c$ is a posterior probability of object class $c$ obtained under the parameter $\theta$, and $C$ is the number of object classes. The entropy $E(x_o)$ is calculated with an image $x_o$ captured by the robot arm. If $E(x_o) > \delta_E$, the classifier requires labeling to the object. This means that the system tries to label the object that the classifier is least certain about its class.

The threshold value is updated as $\delta_E \leftarrow \delta_E + \Delta\delta_E$ after each training of the classifier. $\Delta\delta_E$ is a hyperparameter, and we set it to a minute value. Gradually updating the threshold value to a smaller one during stream-based AL allows us to reduce human annotation costs.

If the classifier requires labeling, the robot arm captures sequential image data $\{x_1,x_2,\cdots,x_{N_{sub}}\} \sim X_{sub}$ while changing the position of its arm tip. Then, the system determines a target image $x^*$ that is with the lowest confidence, in other words, the most effective instance for training the classifier;
\begin{eqnarray}
  x^* = \argmax{}_{x \in \{x_1,x_2,\cdots,x_{N_{sub}}\}} E(x).
  \label{eq:uncertainty}
\end{eqnarray}
Our approach can use effective instances from weakly labeled images $X_{sub}$. That can be considered AL with sub-pooled images.

\setlength\textfloatsep{5pt}
\begin{figure}[t]
  \centering
  \includegraphics[width=8.6cm]{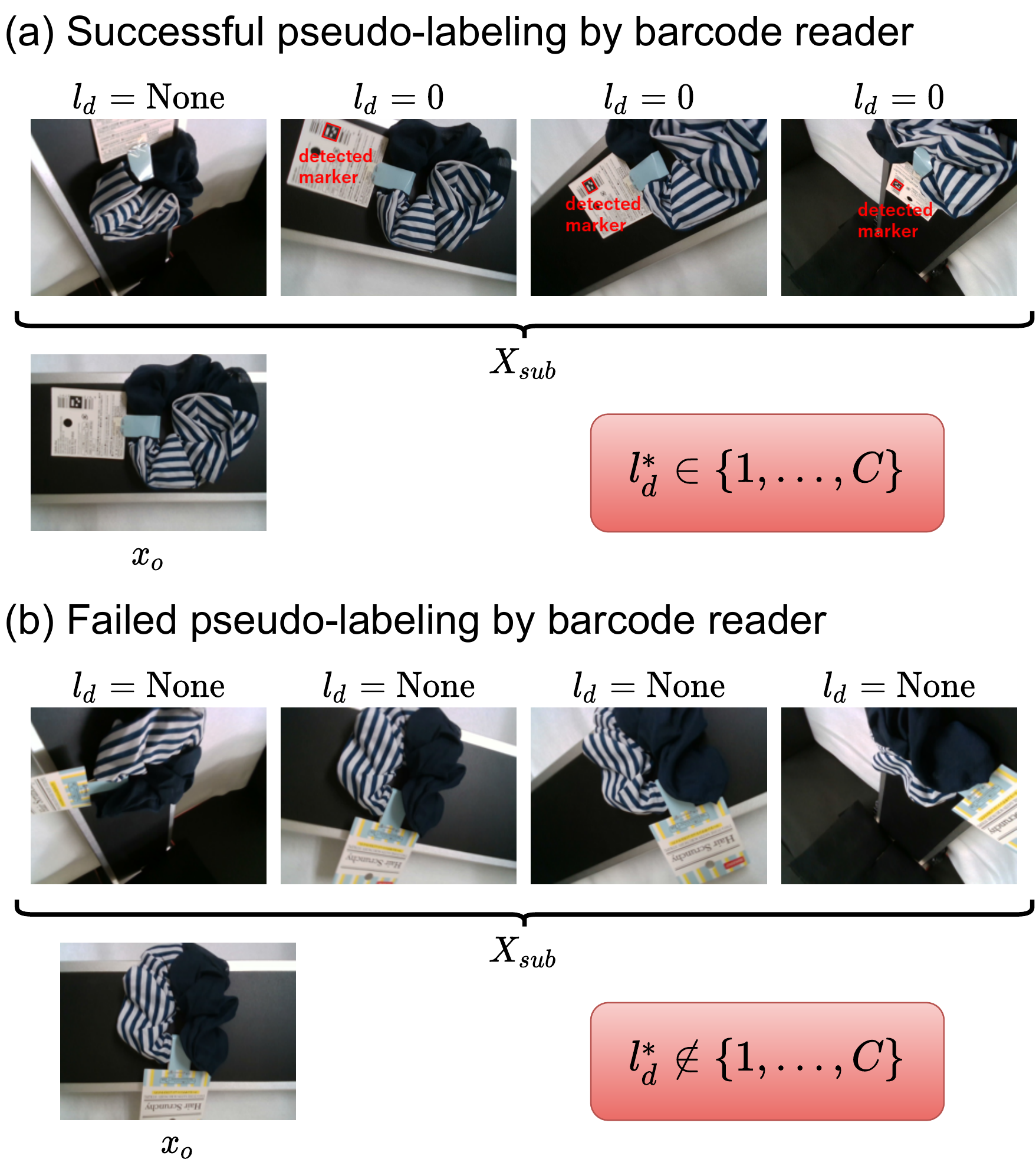}
  \caption{
Successful cases (a) and failed cases (b) of pseudo-labeling by the barcode reader. The most common failed case is when the identifier is hidden behind an object and is undetected.
  }
  \label{fig:label}
\end{figure}

\subsection{Ensemble pseudo-labeling}
\label{sec:psuedo}
If the system decides to add the target image $x^*$ to the training dataset $D$, it adds a label to the images of the object $X_{sub}$ by using the self-training method. Self-training uses the classifier to label images. However, the image $x^*$ that the system requires to label is the most difficult image to label for the classifier. We use images $X_{sub}$ that are captured by the robot arm to address this problem. The object that is in the images $x_o$ and $X_{sub}$ captured by the robot arm are the same objects, thus the labels of objects should be the same. We call images $x_o$ and $X_{sub}$ weakly labeled images based on this fact. Our self-training method, called ensemble pseudo-labeling, has two paths to assign the pseudo-label by using $X_{sub}$: one is by detecting the identifier attached to the object, and the other is via the classifier.

\subsubsection{Pseudo-labeling by Identifier Detector}
First, the system labels images $X_{sub}$ by detecting the identifier~(barcode). We can obtain labels $l_{d,1},l_{d,2},\cdots,l_{d,N_{sub}}$ as labels of the images $X_{sub}$. We use the highest frequency label $l_d^*$ in the obtained labels $l_{d,1},l_{d,2},\cdots,l_{d,N_{sub}}$ as the label of the images $X_{sub}$. Since the system labels the images based on the information obtained by image processing, the system cannot label if the robot arm fails to capture the identifier because of the positions of objects (Fig.~\ref{fig:label}(b)). This problem occurs regardless of hardware.

\subsubsection{Pseudo-labeling by Classifier}
The images that the identifier detector cannot label are labeled with self-training. The classifier assigns pseudo-labels to the images on the basis of 
\begin{eqnarray}
  v = \max{}_{x \in \{x_1,x_2,\cdots,x_{N_{sub}}\}} \hat{f_{\theta}}(x),
  \label{eq:pseudo_label}
\end{eqnarray}
where $\hat{f_{\theta}}(\cdot)$ is the largest value of posterior probabilities obtained using parameter $\theta$. If $v > \delta_v$, the system assigns the pseudo-label $l_p^*$, which is the label of class that gives the value $v$.

These pseudo-labels are obtained from the sequential images $X_{sub}$ separately and allow ensemble pseudo-labeling with high accuracy. By preparing multiple labeling paths with different characteristics, it is possible to complement each other's paths.

\subsection{Human labeling}
\label{sec:human}
If $v \leq \delta_v$ in the process of ensemble pseudo-labeling, we add the image $x^*$ to unlabeled dataset $U$. The proposed method performs human labeling to annotate randomly sampled $x_U^* \sim U$ to label $l_a$ and adds the data ($x_U^*, l_a$) to $D$. This labeling is same as the standard AL method with a human annotator. From the viewpoint of human annotation cost, it is difficult to annotate all images in $U$ every time in updating the classifier in real applications. Thus, we assumed that human labeling is performed to half of the images in $U$ in this study. 

\subsection{Updating the classifier}
\label{sec:update}
The system regularly trains the classifier with pseudo-labeled and human annotated data during the stream-based AL. Based on \cite{semi_m}, we define the loss function $L$ for the classifier as 
\begin{eqnarray}
  L = L_{cls}(f_{\theta}(x), l) + \alpha(n(D)) * L_{cls}(f_{\theta}(x'), l'),
  \label{eq:loss}
\end{eqnarray}
where $l'$ is a pseudo-label assigned by the classifier and $x'$ is its paired data. $l$ is a label assigned by the identifier or human and $x$ is its paired data. $L_{cls}$ is a loss function for the classification, and we use the Softmax cross entropy in this study. Moreover, $\alpha(n(D))$ is a coefficient that controls the effect of the loss function for each kind of label defined as follows:
\begin{eqnarray}
  \alpha(n(D)) = \begin{cases}
  0                        \hspace{40pt} n(D) < T_1, \\
  \frac{n(D)-T_1}{T_2-T_1} \hspace{10pt} T_1 \leq n(D) < T_2, \\
  1                        \hspace{40pt} T_2 \leq n(D),
  \end{cases}
  \label{eq:alpha}
\end{eqnarray}
where $n(D)$ is the number of images in $D$, and $T_1$ and $T_2$ are hyperparameters of $\alpha(n(D))$. $\alpha(n(D))$ is changed by the number of images in $D$. The effect of (pseudo-)labels by the identifier detector or human is larger at the first stage of stream-base AL, and it stabilizes the behavior of stream-based AL. On the other hand, the effect of pseudo-labels by the classifier gradually increases as the training progresses.

\section{Experiment}
\label{sec:exp}

\subsection{Task design and Implementation}
\label{sec:task}
In our experiments, we apply the Robot-Assisted AL (proposed method) to an image-based object classification task and evaluate its effectiveness in terms of the reduction of human annotation cost. The hardware parts of our method were implemented with a robot arm, COBOTTA~\cite{cobotta}, a RealSense D435 camera attached to the hand of the robot arm, and a belt conveyor (see Step 0 in Fig.~\ref{fig:method}). This setup simulates a stream-based AL in a product line in a factory.

\begin{figure}[t]
  \centering
  \includegraphics[width=8.6cm]{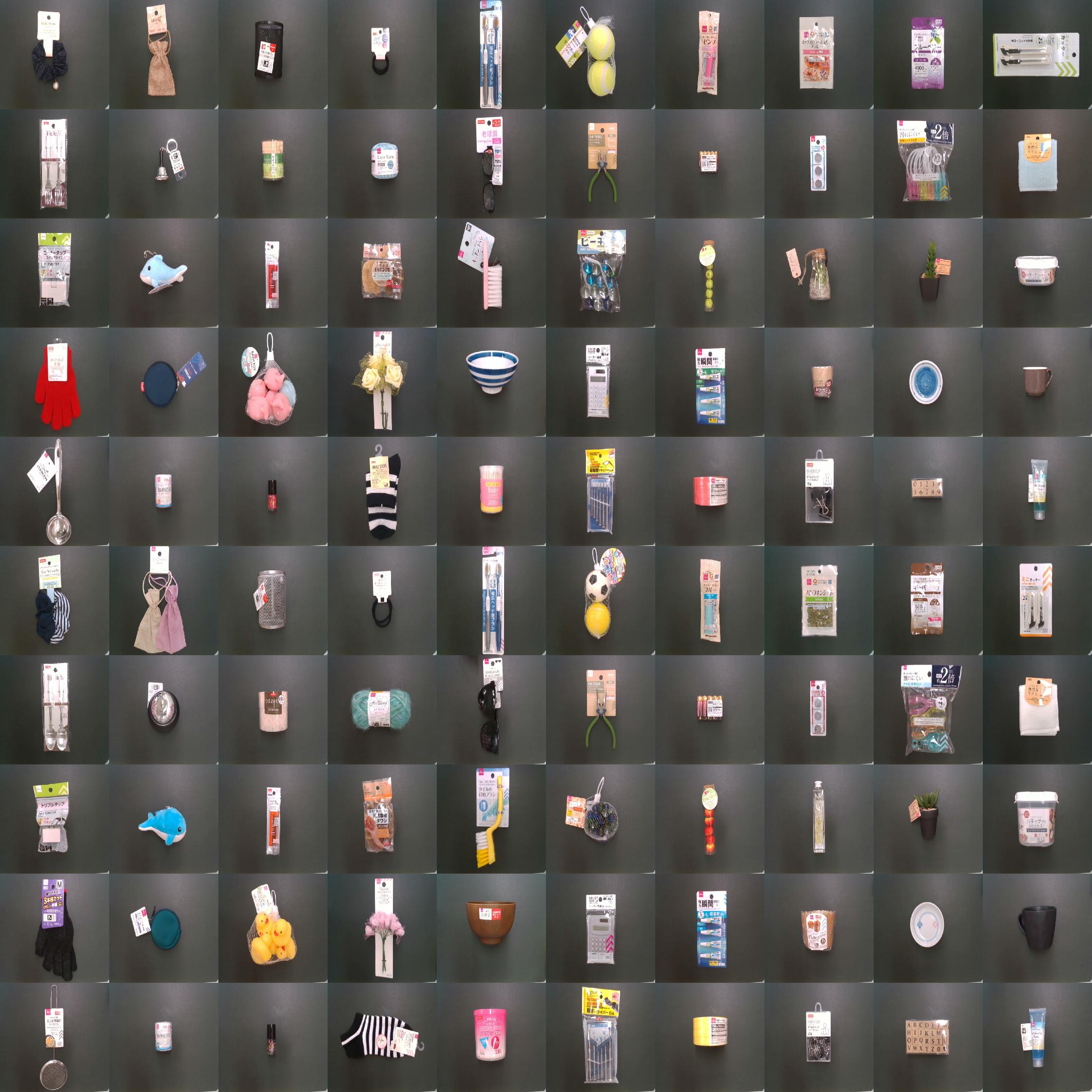}
  \caption{
The objects used for the experiment. We prepared 100 types of miscellaneous goods that can be placed on the belt conveyor. 
  }
  \label{fig:object}
\end{figure}

Fig.~\ref{fig:object} shows the objects used for the experiment. We selected $C=100$ objects of moderate size from various miscellaneous goods so that the hardware system can capture the whole aspect of each object within the arm's movable range. Each object is with an augmented reality (AR) marker coding its class label as a proxy for the commercial barcode,  which is the identifier in this experiment. For the software implementation of the identifier detector, we used a library in OpenCV to detect the AR markers. Each object was placed at a random posture and carried by the belt one by one according to a certain probability distribution. The robot captures an image of an object from above when the object is carried by the belt conveyor and transfers the data to the classifier. If the classifier requires labeling, the robot arm captures sequential images by moving its arm tip about 160 [deg] along the circumference of a radius of about 20 [cm] with respect to the object position (see Step 1 in Fig.~\ref{fig:method}). During the operation, the camera was always aimed at the center of the object. $N_{sub}=80$ images per one object are captured as sequential images $X_{sub}$. We made the dataset obtained in this experiment publicly available~\cite{daiso}.

\setlength\textfloatsep{5pt}

\subsection{Scenarios}
\label{sec:scenario}
\begin{figure}[t]
  \centering
  \includegraphics[width=8.6cm]{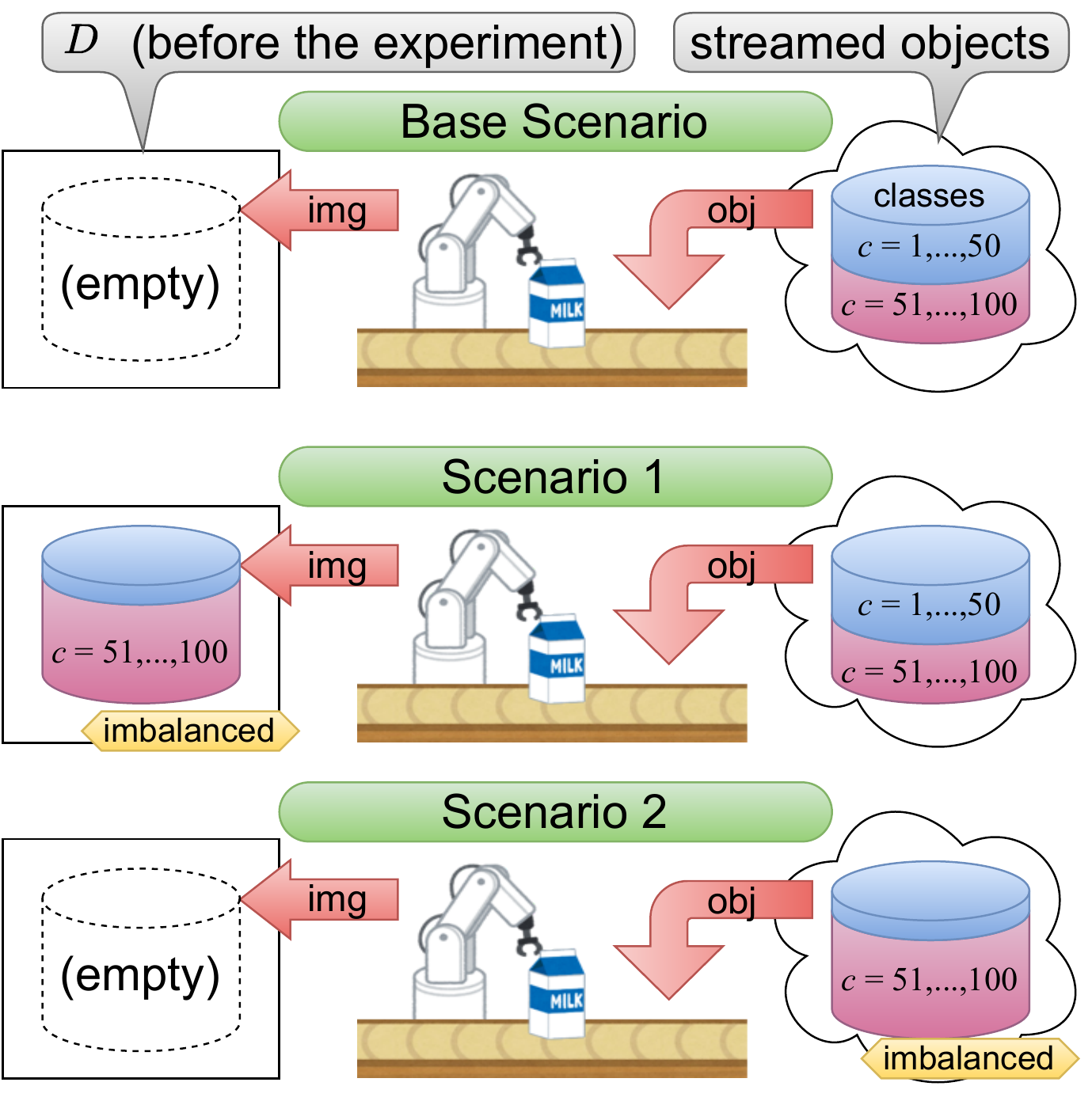}
  \caption{
Three scenarios with different types of temporal variation of data distribution. 
  }
  \label{fig:scenario}
\end{figure}

We evaluate the proposed method in three scenarios of stream-based AL with different types of temporal variation of data distribution (Fig.~\ref{fig:scenario}). The first scenario is the usual setting of stream-based AL~(\textbf{Base Scenario}). In Base Scenario, the initial $D$ is the empty set, and streamed objects on the belt conveyor are uniformly distributed. The other two scenarios are as follows.

\textbf{Scenario 1: } 
The second one is the scenario in which the class distribution in $D$ before the beginning of the experiment is imbalanced. This corresponds to a situation where some products are replaced during the product classification task. In this scenario, there is a certain amount of data of some object classes in $D$ at its initial state. We prepared 30 images for each of the object classes  $c=51,52,\cdots,100$, and added the total $50\times 30=1500$ images to $D$ before the beginning of the experiment. This causes an imbalance between classes in $D$. Other conditions are the same as in the usual stream-based AL. Since the dataset is biased from the beginning of the experiment, the method should resolve it through the stream-based AL in this scenario.

\textbf{Scenario 2: }
The third one is the scenario in which the number of data of each class in $D$ gradually becomes biased if the streamed data are naively added to $D$. This corresponds to a situation where the number of products on the production line is non-uniform. In our experiments, an object flows on the belt conveyor with probability $p$, and this produces the bias. We used an imbalanced class distribution for the stream, that is, the probability $p=1/300$ for object classes $c=1,2,\cdots,50$ and $p=5/300$ for the others.

In Scenarios 1 and 2, it is expected that the bias in the classes affects the performance of the classifier. Scenario 2 is more difficult than Scenario 1, and the method should acquire useful instances more actively.

\subsection{Training setup}
\label{sec:training}
In our experiments, ResNet18~\cite{resnet} was adopted as the image classifier. The data evaluation and the data addition in Algorithm 1 were continued until the data size reaches $N_{max}=4000$. We set $T_1=1000$ and $T_2=2000$ as  hyperparameters of the loss function. The classifier was trained every time $D$ obtains $N_{train}=100$ data. We also set $\delta_E=0.2, \Delta\delta_E=0.01$ and $\delta_v=0.97$. We obtained $\theta'$ by a pre-training with ImageNet. Each training of the classifier was performed with $N_{iter}=100$ iterations. The input to the classifier was $224\times224\times3$ [pixel] RGB image. The batch size was 64. We used random cropping and color augmentations to gain the robustness of the classifier against noise. For the optimization method, stochastic gradient descent with momentum~\cite{sgd} was  employed with 0.01 as the initial learning rate and 0.9 as the momentum. We selected these hyperparameters on the basis of several preliminary trials.

\setlength\textfloatsep{5pt}
\begin{figure*}[t]
    \centering
    \includegraphics[width=17.6cm]{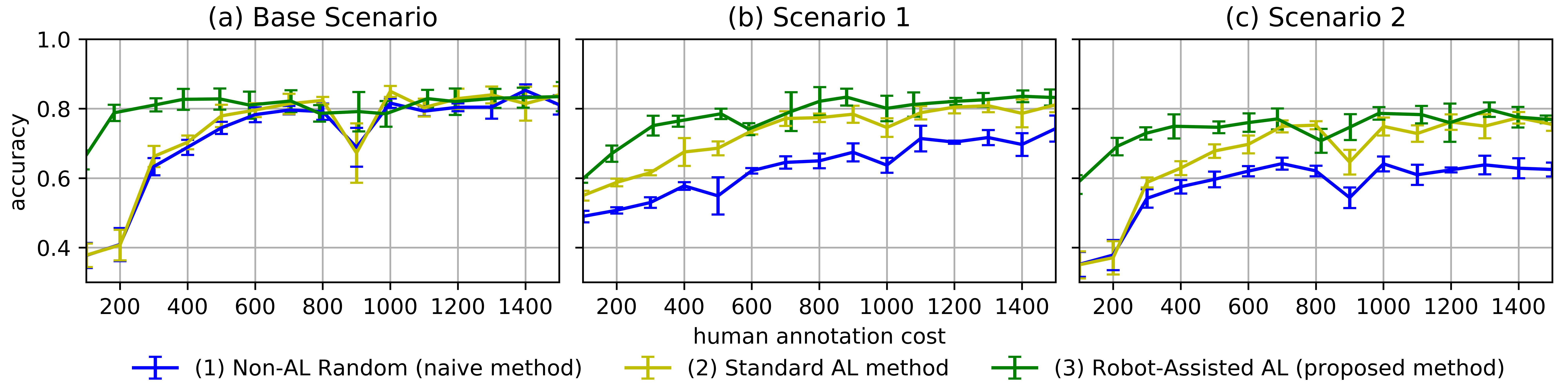}
    \caption{
The test accuracy along with the progress of learning in the usual stream-based AL (Base Scenario), Scenario 1 and Scenario 2. The vertical and horizontal axes correspond to the test accuracy and human annotation cost, respectively. The blue, yellowish-green and green colors correspond to the Non-AL Random method (naive method), Standard AL method and Robot-Assisted AL method (proposed method), respectively. The central lines show the mean accuracy of five trials and the error bars show the standard deviations. All experiments were tried 5 times with different seeds.
    }
    \label{fig:result-acc}
\end{figure*}

\subsection{Performance measurement}
\label{sec:eval}
To evaluate the effectiveness of the proposed method, we compare it with the ``Non-AL Random" method (a naive method)  and the ``Standard AL" method. The Non-AL Random method selects $x^*$ randomly from $X_{sub}$ every time the belt conveyor carries an object. The Standard AL method requires labeling based on the same query strategy described in Sec.~\ref{sec:query} but the $x^*$ is selected randomly from $X_{sub}$. Both the Non-AL Random and Standard AL methods are always given a true label by a human annotator when the classifier requires the label. That is, no data exist in $U$, and only the effect of the query strategy is evaluated purely. These two reference methods assume full human annotation.

We prepared 10 images for each class, 1000 images in total, as the test dataset. Each image was captured by the robot arm from various angles in the same situation as training data. For each experiment, we used five random seeds. The performance of each method is evaluated by the test accuracy along with the annotation cost. The annotation cost is determined by the number of images annotated by a human annotator until the end of the training in the stream-based scenario.

\section{Result and Discussion}\label{SecResult}

\subsection{Performance of the classifier}
We applied the Robot-Assisted Active Learning (proposed method) and the two reference methods to the three stream-based AL scenarios (Fig.~\ref{fig:scenario}). Here, we report the change of test accuracy of the classifier along with the training progress. The test dataset included 1000 images and all experiments were tried 5 times with different seeds (see Sec.~\ref{sec:eval}).

\textbf{Base Scenario: }
Fig.~\ref{fig:result-acc}(a) shows the results of the three methods applied to the usual stream-based AL scenario (Base Scenario). Each line shows the change of test accuracy of each method as the number of human-annotated data increases. We can see that the Robot-Assisted AL (3) achieved coequal performance with the other two methods (1,2) with fewer amount of human annotations. The Robot-Assisted AL reduced about 60\% of the human annotation cost until the convergence of the learning. This clearly shows the effectiveness of the self-training with ensemble pseudo-labeling in the Robot-Assisted AL. The same trend was also observed in Scenarios 1 and 2. On the other hand, the Non-AL Random exhibited coequal accuracy with the Standard AL in this scenario where the dataset contains uniformly distributed classes. This suggests that the effect of AL is insignificant in long term in such situations. 

\textbf{Scenario 1: } 
In this scenario, we compare the behaviors of the three methods when the class distribution in $D$ before the beginning of the experiment is imbalanced. Fig.~\ref{fig:result-acc}(b) shows the results. The AL-based methods (2,3) kept improving the accuracy from the early stage of learning and reached the accuracy coequal to the ones achieved in Base Scenario in the end. In particular, the Robot-Assisted AL (3) had the best accuracy among the three methods, which indicated that the Robot-Assisted AL collects effective data efficiently. On the other hand, although the Non-AL Random (1) exhibits a low accuracy in the early stage of learning due to the bias in $D$, the accuracy improved in the later stage. If the learning period is further extended, it is expected that the accuracy of the Non-AL Random will approach the accuracy of the AL-based methods because the amount of data will be so large that the bias in $D$ at the initial state will be negligible. 

\textbf{Scenario 2: }
In this scenario, we compare the behaviors of the three methods in the case where the streamed data is biased. Fig.~\ref{fig:result-acc}(c) shows the change of accuracy of each method in this scenario as the number of human-annotated data increases. The Non-AL Random (1) showed lower accuracy than in Scenario 1, while the AL-based methods, including the Robot-Assisted AL, achieved high accuracy as in Base Scenario and Scenario 1. This is because the Non-AL Random could not resolve the bias in data, even if it collects a large amount, unlike in Scenario 1. The AL-based methods are most effective in this scenario.

We have confirmed that the Robot-Assisted AL collects effective data more efficiently and reduces a larger amount of human annotation cost compared to the other two methods in all scenarios. In the next subsection, we investigate the data distributions in the training dataset obtained during stream-based AL to understand the detailed contributions of AL.

\setlength\textfloatsep{5pt}
\begin{figure*}[t]
    \centering
    \includegraphics[width=17.6cm]{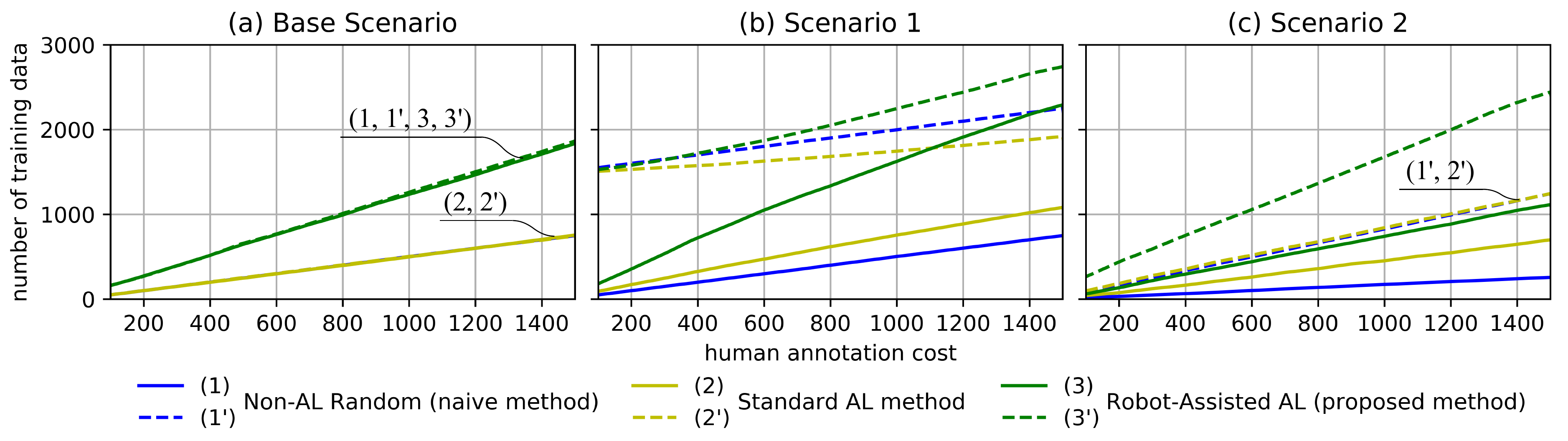}
    \caption{
The number of training data $n(D)$ along with the progress of learning in the usual stream-based AL scenario (Base Scenario), Scenario 1 and Scenario 2. The vertical axis is the number of training data, and the horizontal axis is the human annotation cost. The solid and dashed lines show the mean number of data corresponding to the object classes  $c=1,2,\cdots,50$ and $c=51,52,\cdots,100$ calculated with five trials, respectively. The blue, yellowish-green and green lines correspond to the Non-AL Random method (naive method), Standard AL method and the Robot-Assisted AL method (proposed method), respectively. Error bars are omitted in this figure to keep readability. All experiments were tried 5 times with different seeds.
    }
    \label{fig:result-data}
\end{figure*}

\begin{table*}[t]
    \centering
    \begin{tabular}{c|ccc}
        \multicolumn{4}{c}{TABLE I: Pseudo-labeled data by classifier} \\
        \hline
                     & Base Scenario    & Scenario 1       & Scenario 2 \\
        \hline \hline
        Accuracy of $l_p^*$             & 98.9$\pm{0.8}\%$ & 97.6$\pm{1.3}\%$ & 97.0$\pm{2.7}\%$ \\ 
        \hline
        Ratio of ($x^*$, $l_p^*$) in $D$  & 2.9$\pm{0.3}\%$  &  2.3$\pm{0.1}\%$ &  2.0$\pm{0.7}\%$ \\ 
        \hline 
    \end{tabular}
\end{table*}

\subsection{Collected training dataset}
Fig.~\ref{fig:result-data} shows the numbers of training data that have labels 1--50 and the numbers of training data that have labels 51--100 obtained during stream-based AL in each scenario. The solid and dashed lines show the number of data corresponding to the object classes $c=1,2,\cdots,50$ and  $c=51,52,\cdots,100$, respectively. 

In Base Scenario, where the class distribution is uniform for the stream, the solid and the dashed lines overlap, and there are no differences between the classes  (Fig.~\ref{fig:result-data}(a)). The inclination of the lines corresponding to the Robot-Assisted AL (3) is steeper than those corresponding to the other two methods (1,2), which means that the Robot-Assisted AL collected data efficiently with fewer  human annotations.

In Scenario 1, the solid lines and the dashed lines split  due to the bias that exists between the data corresponding to the object classes $c=1,2,\cdots,50$ and the data corresponding to $c=51,52,\cdots,100$ (Fig.~\ref{fig:result-data}(b)). The lines corresponding to the Non-AL Random (1) keep the distance. On the other hand, the solid and dashed lines corresponding to the AL-based methods (2,3) gradually get closer to each other. This is because AL-based methods try to resolve the bias that exists between the classes. In particular, the solid and dashed lines corresponding to the Robot-Assisted AL (3) get closer than those corresponding to the Standard AL (2), which means that the Robot-Assisted AL resolves more bias with fewer amounts of annotations.

In Scenario 2, the solid and  dashed lines split apart due to the imbalanced class distribution of the streamed objects (Fig.~\ref{fig:result-data}(c)). In this scenario, the distances between the solid and dashed lines gradually become larger for all methods. The degree of this increment in distance corresponding to the AL-based methods (2,3) is smaller than in the Non-AL Random (1). This is because the AL-based methods (2,3) work to intensively collect less streamed data. The distance between the solid and dashed lines corresponding to the Robot-Assisted AL (3) is the widest among all. This is due to the high efficiency of data collection per human annotation.

\subsection{Pseudo-labels}
Table I shows the accuracy of the pseudo-label by the classifier of the proposed method (first row), and the ratio of the pseudo-labeled data by the classifier in the training dataset (second row). The accuracies are higher than 97\% in all scenarios, and the ratios of the pseudo-labels in $D$ range within about 2--3\%, which means that the pseudo-labels given by the classifier slightly contributed to reducing  human annotation cost. Although the contribution of the pseudo-labels assigned by the classifier to the learning is smaller than that by the identifier detector, the absolute amount of the human annotation cost reduction is likely to increase to a non-negligible level in long-term learning.

As described in Section~\ref{sec:intro}, the self-training highly depends on the classifier's prediction accuracy, and thus it usually does not work stably in the stream-based AL. If the system cannot use the pseudo-labels given by the identifier detector nor cannot use weakly labeled data, then providing high confidence data to the classifier is difficult. The long-term learning process may become unstable in such cases.

Clearly, the experiment results show the effectiveness of the proposed method. In the next section, we discuss limitations of this study.

\section{Limitation}
The main limitation of this study is threefold. First, the proposed method does not achieve complete elimination of human annotation cost. It requires human annotation for half of the training images, other than the ones given the pseudo-labels. In our pre-experiments, we have tried adversarial training~\cite{semi1} and metric learning~\cite{semi2} to the stream-based AL, but the accuracy did not improve. Although these methods are promising approaches to reduce annotation cost, the effective situations are limited. The previous work~\cite{semi5} shows the effectiveness of self-training without robot assistance in a certain situation of AL. We plan to improve the self-training algorithm itself for stream-based AL referencing \cite{semi3,semi4}.

Second, the test accuracies reported in Sec.~\ref{SecResult} were measured by test sets that have the same class distributions as those of the streamed objects. The behavior of the proposed method may affect test accuracy differently if the distribution of the test set is different. From this viewpoint, we need further investigations.

Third, we have tested the proposed algorithm only with AR marker in this study, which is a highly standardized identifier. We can also consider other types of identifiers like the ones listed in Subsec.~\ref{sec:AL}. Although the experiments in this study show the validity of the core concept of the proposed method, we need experiments for other types of identifiers when we deploy the proposed method to real-world applications. In addition, the camera angles to capture images and find the identifier are human-designed in this study. It is possible to estimate an efficient camera angle depending on the object pose or change the pose by moving the object by the robot arm. Extending the proposed method by adding these functions is another direction of future work.

\section{Conclusion}
In this study, we have proposed a stream-based active learning method assisted by a robot that reduces human annotation cost. The proposed method allows pseudo-labeling with high accuracy by using sequential images captured by a robot arm, that are used as weakly labeled data. We applied the proposed method to classify objects that are carried by a belt conveyor in three scenarios with various data distributions. The results show that the proposed method can achieve the same accuracy with half labeling costs compared with the reference methods without the robot assistance. In future work, we plan to extend our method to handle the case in which high-confidence identifiers such as the AR marker are not available.

\section*{Acknowledgment}
This work was supported by JST, ACT-X Grant Number JPMJAX190I, Japan.

\bibliographystyle{IEEEtran}
\bibliography{refs}

\begin{thebibliography}{10}
\providecommand{\url}[1]{#1}
\csname url@samestyle\endcsname
\providecommand{\newblock}{\relax}
\providecommand{\bibinfo}[2]{#2}
\providecommand{\BIBentrySTDinterwordspacing}{\spaceskip=0pt\relax}
\providecommand{\BIBentryALTinterwordstretchfactor}{4}
\providecommand{\BIBentryALTinterwordspacing}{\spaceskip=\fontdimen2\font plus
\BIBentryALTinterwordstretchfactor\fontdimen3\font minus
  \fontdimen4\font\relax}
\providecommand{\BIBforeignlanguage}[2]{{%
\expandafter\ifx\csname l@#1\endcsname\relax
\typeout{** WARNING: IEEEtran.bst: No hyphenation pattern has been}%
\typeout{** loaded for the language `#1'. Using the pattern for}%
\typeout{** the default language instead.}%
\else
\language=\csname l@#1\endcsname
\fi
#2}}
\providecommand{\BIBdecl}{\relax}
\BIBdecl

\bibitem{dl1}
A.~Krizhevsky, I.~Sutskever, and G.~E. Hinton, ``Imagenet classification with
  deep convolutional neural networks,'' in \emph{Advances in Neural Information
  Processing Systems}, vol.~25, 2012.

\bibitem{dl2}
Z.~Alom, T.~M. Taha, C.~Yakopcic, S.~Westberg, P.~Sidike, M.~S. Nasrin,
  M.~Hasan, B.~C.~V. Essen, A.~A.~S. Awwal, and V.~K. Asari, ``A
  state-of-the-art survey on deep learning theory and architectures,''
  \emph{Electronics}, vol.~8, no.~3, p. 292, 2019.

\bibitem{an1}
H.~Su, J.~Deng, and L.~Fei-Fei, ``Crowdsourcing annotations for visual object
  detection.'' in \emph{AAAI 4th Human Comput. Workshop}, 2012.

\bibitem{an2}
P.~D. Papadopoulos, R.~R.~J. Uijlings, F.~Keller, and V.~Ferrari, ``We don't
  need no bounding-boxes: Training object class detectors using only human
  verification,'' in \emph{IEEE/CVF Conf. on Computer Vision and Patt. Recog.},
  2016.

\bibitem{an3}
K.~Konyushkova, J.~Uijlings, C.~H. Lampert, and V.~Ferrari, ``Learning
  intelligent dialogs for bounding box annotation,'' in \emph{IEEE/CVF Conf. on
  Computer Vision and Patt. Recog.}, 2018.

\bibitem{online1}
S.~{Rebuffi}, A.~{Kolesnikov}, G.~{Sperl}, and C.~H. {Lampert}, ``icarl:
  Incremental classifier and representation learning,'' in \emph{IEEE/CVF Conf.
  on Computer Vision and Patt. Recog.}, 2017.

\bibitem{online2}
V.~Lomonaco and D.~Maltoni, ``Core50: a new dataset and benchmark for
  continuous object recognition,'' in \emph{Annual Conf. on Robot Learning},
  2017.

\bibitem{online3}
T.~Xiao, J.~Zhang, K.~Yang, Y.~Peng, and Z.~Zhang, ``Error-driven incremental
  learning in deep convolutional neural network for large-scale image
  classification,'' in \emph{ACM Int. Conf. on Multimedia}, 2014.

\bibitem{ent}
B.~Settles, ``Active learning literature survey,'' University of
  Wisconsin--Madison, Computer Sciences Technical Report 1648, 2009.

\bibitem{al2}
S.~Sinha, S.~Ebrahimi, and T.~Darrell, ``Variational adversarial active
  learning,'' in \emph{IEEE Conf. on Computer Vision and Patt. Recog.}, 2019.

\bibitem{al3}
D.~{Yoo} and I.~S. {Kweon}, ``Learning loss for active learning,'' in
  \emph{IEEE/CVF Conf. on Computer Vision and Patt. Recog.}, 2019.

\bibitem{al4}
C.~Mayer and R.~Timofte, ``Adversarial sampling for active learning,''
  \emph{arXiv preprint arXiv:1808.06671}, 2018.

\bibitem{semi1}
T.~Miyato, S.~Maeda, M.~Koyama, K.~Nakae, and S.~Ishii, ``Distributional
  smoothing with virtual adversarial training,'' \emph{Int. Conf. on Learning
  Representations}, 2016.

\bibitem{semia2}
M.~{Guillaumin}, J.~{Verbeek}, and C.~{Schmid}, ``Multimodal semi-supervised
  learning for image classification,'' in \emph{IEEE/CVF Conf. on Computer
  Vision and Patt. Recog.}, 2010.

\bibitem{semi2}
E.~Hoffer and N.~Ailon, ``Deep metric learning using triplet network,'' in
  \emph{Int. Workshop on Similarity-Based Pattern Recognition}, 2015.

\bibitem{rep1}
M.~Tschannen, O.~Bachem, and M.~Lucic, ``Recent advances in autoencoder-based
  representation learning,'' \emph{arXiv preprint arXiv:1812.05069}, 2018.

\bibitem{domein1}
S.~Motiian, Q.~Jones, S.~Iranmanesh, and G.~Doretto, ``Few-shot adversarial
  domain adaptation,'' in \emph{Advances in Neural Information Processing
  Systems}, vol.~30, 2017.

\bibitem{trans1}
Y.~Ganin, E.~Ustinova, H.~Ajakan, P.~Germain, H.~Larochelle, F.~Laviolette,
  M.~Marchand, and V.~Lempitsky, ``Domain-adversarial training of neural
  networks,'' vol.~17, no.~1, pp. 2030--2096, 2016.

\bibitem{co1}
A.~Blum and T.~Mitchell, ``Combining labeled and unlabeled data with
  co-training,'' in \emph{Annual Conf. on Computational Learning Theory}, 1998.

\bibitem{co2}
S.~Qiao, W.~Shen, Z.~Zhang, B.~Wang, and A.~Yuille, ``Deep co-training for
  semi-supervised image recognition,'' 2018.

\bibitem{co3}
K.~Saito, Y.~Ushiku, and T.~Harada, ``Asymmetric tri-training for unsupervised
  domain adaptation,'' in \emph{Int. Conf. on Machine Learning}, 2017.

\bibitem{r1}
L.~Pinto and A.~Gupta, ``Supersizing self-supervision: Learning to grasp from
  50k tries and 700 robot hours,'' 2015.

\bibitem{r2}
K.~Suzuki, Y.~Yokota, Y.~Kanazawa, and T.~Takebayashi, ``Online self-supervised
  learning for object picking: Detecting optimum grasping position using a
  metric learning approach,'' in \emph{IEEE/SICE Int. Symp. on System
  Integrations}, 2020.

\bibitem{r3}
Y.~Yokota, K.~Suzuki, Y.~Kanazawa, and T.~Takebayashi, ``A multi-task learning
  framework for grasping-position detection and few-shot classification,'' in
  \emph{IEEE/SICE Int. Symp. on System Integrations}, 2020.

\bibitem{semi_m}
D.~hyun Lee, ``Pseudo-label: The simple and efficient semi-supervised learning
  method for deep neural networks,'' in \emph{ICML Workshop on Challenges in
  Representation Learning}, 2013.

\bibitem{cobotta}
{DENSO Wave}, ``{COBOTTA},'' available on:
  \url{https://www.denso-wave.com/en/robot/product/collabo/cobotta.html}.

\bibitem{daiso}
{Fujitsu Limited}, ``{DAISO-100},'' available on:
  \url{http://dataset.jp.fujitsu.com/data/daiso100/index.html}.

\bibitem{resnet}
K.~{He}, X.~{Zhang}, S.~{Ren}, and J.~{Sun}, ``Deep residual learning for image
  recognition,'' in \emph{IEEE/CVF Conf. on Computer Vision and Patt. Recog.},
  2016.

\bibitem{sgd}
I.~Sutskever, J.~Martens, G.~Dahl, and G.~Hinton, ``On the importance of
  initialization and momentum in deep learning,'' in \emph{Int. Conf. on
  Machine Learning}, vol.~28, 2013.

\bibitem{semi5}
O.~Siméoni, M.~Budnik, Y.~Avrithis, and G.~Gravier, ``Rethinking deep active
  learning: Using unlabeled data at model training,'' \emph{arXiv preprint
  arXiv:1911.08177}, 2019.

\bibitem{semi3}
I.~Muslea, S.~Minton, and C.~A. Knoblock, ``Active + semi-supervised learning =
  robust multi-view learning,'' in \emph{Int. Conf. on Machine Learning}, 2002.

\bibitem{semi4}
X.~Zhu, J.~Lafferty, and Z.~Ghahramani, ``Combining active learning and
  semi-supervised learning using gaussian fields and harmonic functions,'' in
  \emph{ICML Workshop on The Continuum from Labeled to Unlabeled Data in
  Machine Learning and Data Mining}, 2003.

\end{thebibliography}

\end{document}